\documentclass[journal]{IEEEtran}

\usepackage{fancyvrb}
\usepackage[pdftex]{graphicx}
\usepackage{hyperref}
\usepackage{ragged2e}
\usepackage[section]{placeins}
\usepackage{cite}

\begin{document}

\title{Big Data and the SP Theory of Intelligence}

\author{J Gerard Wolff,~\IEEEmembership{Member,~IEEE}%
\thanks{Gerry Wolff is founder and director of CognitionResearch.org, Menai Bridge, UK. e-mail: jgw@cognitionresearch.org.}%
\thanks{Manuscript received [Month] [day], [year]; revised [Month] [day], [year].}}

\markboth{IEEE Access,~Vol.~X, No.~Y, [month]~[year]}%
{Wolff: Big Data and the SP Theory of Intelligence}

\maketitle

\begin{abstract}

This article is about how the {\em SP theory of intelligence} and its realisation in the {\em SP machine} may, with advantage, be applied to the management and analysis of big data. The SP system---introduced in the article and fully described elsewhere---may help to overcome the problem of variety in big data: it has potential as {\em a universal framework for the representation and processing of diverse kinds of knowledge} (UFK), helping to reduce the diversity of formalisms and formats for knowledge and the different ways in which they are processed. It has strengths in the unsupervised learning or discovery of structure in data, in pattern recognition, in the parsing and production of natural language, in several kinds of reasoning, and more. It lends itself to the analysis of streaming data, helping to overcome the problem of velocity in big data. Central in the workings of the system is lossless compression of information: making big data smaller and reducing problems of storage and management. There is potential for substantial economies in the transmission of data, for big cuts in the use of energy in computing, for faster processing, and for smaller and lighter computers. The system provides a handle on the problem of veracity in big data, with potential to assist in the management of errors and uncertainties in data. It lends itself to the visualisation of knowledge structures and inferential processes. A high-parallel, open-source version of the SP machine would provide a means for researchers everywhere to explore what can be done with the system and to create new versions of it.

\end{abstract}


\section{Introduction}\label{introduction_section}

Big data---the large volumes of data that are now produced in many fields---can present problems in storage, transmission, and processing, but their analysis may yield useful information and useful insights.

This article is about how the {\em SP theory of intelligence} and its realisation in the {\em SP machine} (Section \ref{introduction_to_sp_section}) may, with advantage, be applied to big data. Naturally, in an area like that, problems will not be solved in one step. The ideas described in this article provide a foundation and framework for further research (Section \ref{road_map_section}).

Problems associated with big data are reviewed quite fully in {\em Frontiers in Massive Data Analysis} \cite{national_research_council_2013} from the US National Research Council, and there is another useful perspective, from IBM, in {\em Smart Machines: IBM's Watson and the Era of Cognitive Computing} \cite{kelly_hamm_2013}. These and other sources are referenced at appropriate points throughout the article.

In broad terms, the potential benefits of the SP system, as applied to big data, are in these areas:

\begin{itemize}

\item {\em Overcoming the problem of variety in big data}. Harmonising diverse kinds of knowledge, diverse formats for knowledge, and their diverse modes of processing, via a universal framework for the representation and processing of knowledge.

\item {\em Learning and discovery}. The unsupervised learning or discovery of `natural' structures in data.

\item {\em Interpretation of data}. The SP system has strengths in areas such as pattern recognition, information retrieval, parsing and production of natural language, translation from one representation to another, several kinds of reasoning, planning and problem solving.

\item {\em Velocity: analysis of streaming data}. The SP system lends itself to an incremental style, assimilating information as it is received, much as people do.

\item {\em Volume: making big data smaller}. Reducing the size of big data via lossless compression can yield direct benefits in the storage, management, and transmission of data, and indirect benefits in several of the other areas discussed in this article.

\item {\em Additional economies in the transmission of data}. There is potential for additional economies in the transmission of data, potentially very substantial, by judicious separation of `encoding' and `grammar'.

\item {\em Energy, speed, and bulk}. There is potential for big cuts in the use of energy in computing, for greater speed of processing with a given computational resource, and for corresponding reductions in the size and weight of computers.

\item {\em Veracity: managing errors and uncertainties in data}. The SP system can identify possible errors or uncertainties in data, suggest possible corrections or interpolations, and calculate associated probabilities.

\item {\em Visualisation}. Knowledge structures created by the system, and inferential processes in the system, are all transparent and open to inspection. They lend themselves to display with static and moving images.

\end{itemize}

These topics will be discussed, each in its own section, below. But first, the SP theory and the SP machine will be introduced.

\section{Introduction to the SP Theory and SP Machine}\label{introduction_to_sp_section}

The SP theory, which has been under development for several years, aims to simplify and integrate concepts across artificial intelligence, mainstream computing and human perception and cognition, with information compression as a unifying theme.

\begin{figure}[!htbp]
\centering
\includegraphics[width=0.4\textwidth]{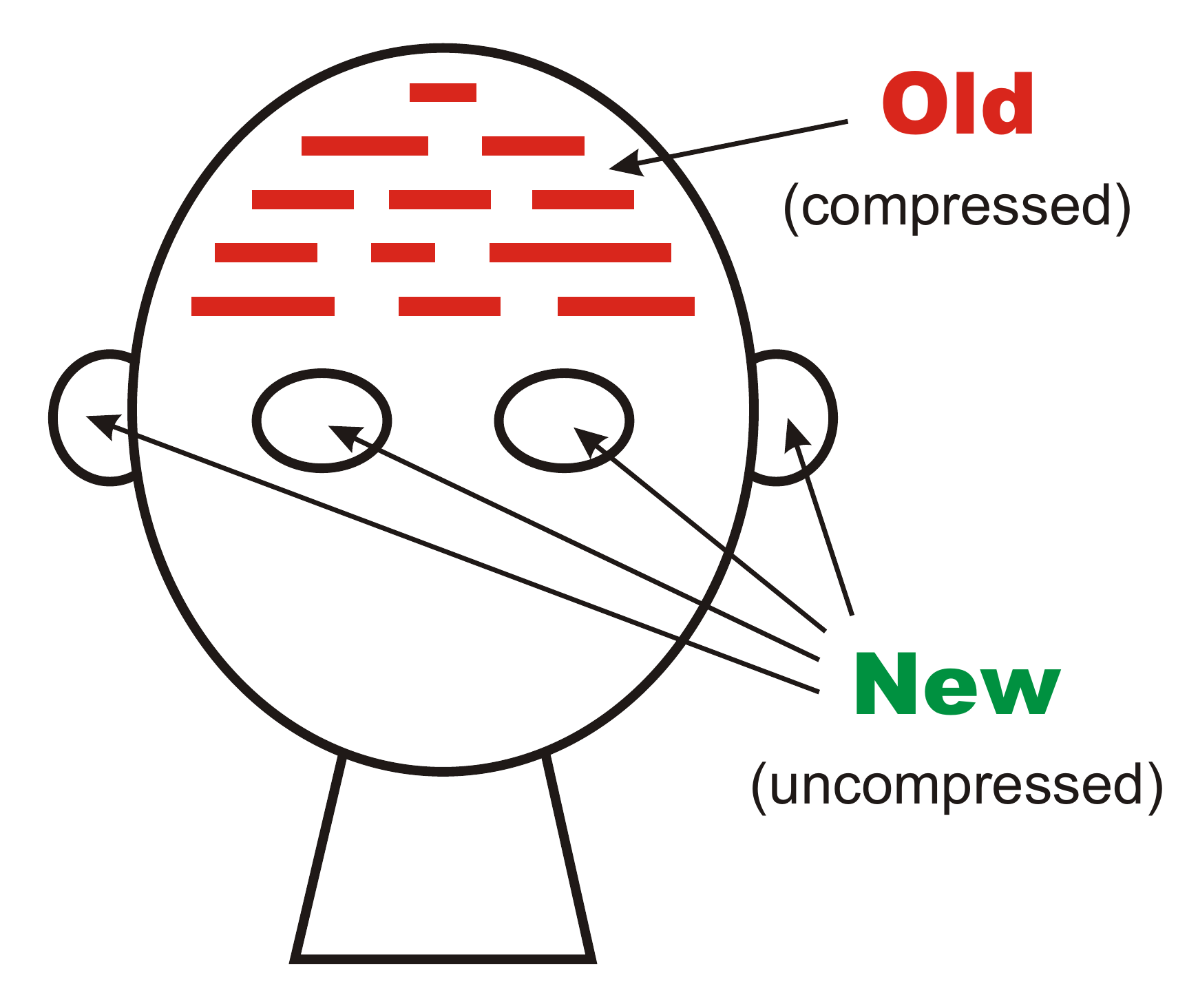}
\caption{Schematic representation of the SP system from an `input' perspective. Reproduced from Fig.~1 in \cite{sp_extended_overview}, with permission.}
\label{sp_input_perspective_figure}
\end{figure}

The theory is conceived as an abstract brain-like system that, in an `input' perspective, may receive {\em New} information via its senses, and compress some or all of it to create {\em Old} information, as illustrated schematically in Fig.~\ref{sp_input_perspective_figure}. In the theory, information compression is the mechanism both for the learning and organisation of knowledge and for pattern recognition, reasoning, problem solving, and more.

In the SP system, all kinds of knowledge are represented with {\em patterns}: arrays of atomic symbols in one or two dimensions.

\begin{figure*}[!htbp]
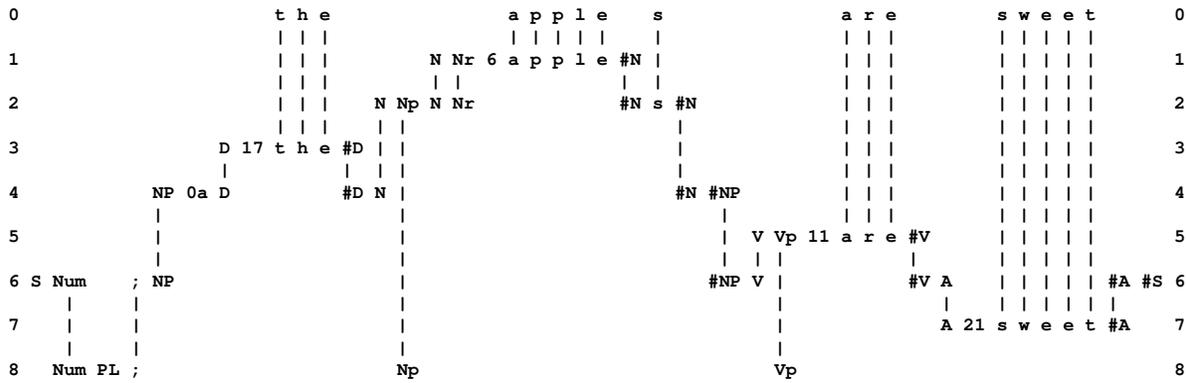

\fontsize{07.00pt}{08.40pt}
\centering
{\bf
\begin{BVerbatim}
0                       t h e                a p p l e    s                a r e         s w e e t       0
                        | | |                | | | | |    |                | | |         | | | | |
1                       | | |         N Nr 6 a p p l e #N |                | | |         | | | | |       1
                        | | |         | |              |  |                | | |         | | | | |
2                       | | |    N Np N Nr             #N s #N             | | |         | | | | |       2
                        | | |    | |                        |              | | |         | | | | |
3                  D 17 t h e #D | |                        |              | | |         | | | | |       3
                   |          |  | |                        |              | | |         | | | | |
4            NP 0a D          #D N |                        #N #NP         | | |         | | | | |       4
             |                     |                            |          | | |         | | | | |
5            |                     |                            |  V Vp 11 a r e #V      | | | | |       5
             |                     |                            |  | |           |       | | | | |
6 S Num    ; NP                    |                           #NP V |           #V A    | | | | | #A #S 6
     |     |                       |                                 |              |    | | | | | |
7    |     |                       |                                 |              A 21 s w e e t #A    7
     |     |                       |                                 |
8   Num PL ;                       Np                                Vp                                  8
\end{BVerbatim}
}
\caption{A multiple alignment created by the SP computer model that achieves the effect of parsing a sentence (`\texttt{t h e a p p l e s a r e s w e e t}'). Reproduced from Fig.~1 in \cite{sp_benefits_apps}, with permission.}
\label{parsing_1_figure}
\end{figure*}

At the heart of the system are processes for compressing information by finding good full and partial matches between patterns and merging or `unifying' parts that are the same. More specifically, all processing is done via the creation of {\em multiple alignments}, like the one shown in Fig.~\ref{parsing_1_figure}.\footnote{\raggedright The concept of multiple alignment in the SP system \cite[Section 4]{sp_extended_overview}, \cite[Section 3.4]{wolff_2006} is borrowed from that concept in bioinformatics, but with important differences.}

The close association between information compression and concepts of prediction and probability \cite{li_vitanyi_2009} means that the SP system is intrinsically probabilistic. Each SP pattern has an associated frequency of occurrence, and for each multiple alignment, the system may calculate associated probabilities \cite[Section 3.7]{wolff_2006} (reproduced in \cite[Section 4.4]{sp_extended_overview}). Although the SP system is fundamentally probabilistic, it can, if required, be constrained to operate in the clockwork style of a conventional computer, delivering all-or-nothing results \cite[Chapter 10]{wolff_2006}.

An important idea in the SP programme is the {\em DONSVIC} principle \cite[Section 5.2]{sp_extended_overview}: the conjecture, supported by evidence, that information compression, properly applied, is the key to the discovery of `natural' structures, meaning the kinds of things that people naturally recognise, such as words, objects, and classes of objects. Evidence to date suggests that the SP system does indeed conform to that principle.

The SP theory is realised in a computer model, SP70, which may be regarded as a first version of the SP machine. It is envisaged that the SP computer model will provide the basis for the development of a high-parallel, open-source version of the SP machine, as described in Section \ref{road_map_section}.

The theory has things to say about several aspects of computing and cognition, including unsupervised learning, concepts of computing, aspects of mathematics and logic, the representation of knowledge, natural language processing, pattern recognition, several kinds of reasoning, information storage and retrieval, planning and problem solving, and aspects of neuroscience and of human perception and cognition.

\sloppy There is a relatively full account of the SP system in \cite{wolff_2006}, an extended overview in \cite{sp_extended_overview}, an account of its existing and expected benefits and applications in \cite{sp_benefits_apps}, a description of its foundations in \cite{sp_foundations}, and an introduction to the system in \cite{sp_introduction}. More information may be found via \href{http://www.cognitionresearch.org/sp.htm}{www.cognitionresearch.org/sp.htm}.

\section{Overcoming the Problem of Variety in Big Data}\label{problem_of_variety_section}

\begin{quote}

``The manipulation and integration of heterogeneous data from different sources into a meaningful common representation is a major challenge.'' \cite[p.~76]{national_research_council_2013}.

\end{quote}

\begin{quote}

``Over the past decade or so, computer scientists and mathematicians have become quite proficient at handling specific types of data by using specialized tools that do one thing very well.~...~But that approach doesn't work for complex operational challenges such as managing cities, global supply chains, or power grids, where many interdependencies exist and many different kinds of data have to be taken into consideration.'' \cite[p.~48]{kelly_hamm_2013}.

\end{quote}

The many different kinds of data include: the world's many languages, spoken or written; static and moving images; music as sound and music in its written form; numbers and mathematical notations; tables; charts; graphs; networks; trees; grammars; computer programs; and more. With many of these kinds of data, there are several different computer-based formats, such as, with static images: JPEG, TIFF, WMF, BMP, GIF, EPS, PDF, PNG, PBM, and more. And, normally, each kind of data, and each different format, needs to be processed in its own special way.

Some of this diversity is necessary and useful. For example:

\begin{itemize}

\item The cultural life of a community is often intimately connected with the language of that community.

\item Notwithstanding the dictum that ``A picture is worth a thousand words'', natural languages, collectively, have special strengths.

\item Ancient texts are of interest for historical, cultural and other reasons.

\item With techniques and technologies as they have developed to date, it often makes sense to use different formalisms or formats for different purposes.

\item Over-zealous standardisation may stifle creativity.

\end{itemize}

Nevertheless, there are several reasons, described in the next subsection, for trying to develop {\em a universal framework for the representation and processing of diverse kinds of knowledge} (UFK). Such a system may help to reduce unnecessary diversity in formalisms and formats for knowledge and in their modes of processing. But it is likely that many existing systems would continue in use for the kinds of reasons mentioned above, perhaps with translations into UFK form, if or when that proves necessary.

\subsection{Reasons for Developing a Universal Framework for the Representation and Processing of Knowledge}

Of the reasons described here for developing a UFK, some relate fairly directly to issues with big data (Sections \ref{variety_discovery_section}, \ref{variety_interpretation_section}, and \ref{variety_nl_section}), while the rest draw on other aspects of computing, engineering, and biology.

\subsubsection{The Discovery of Structure in Data}\label{variety_discovery_section}

If we are trying to discover patterns of association or other structures in big data (Section \ref{learning_discovery_section}), a diversity of formalisms and formats is a handicap. Let us imagine for example how an artificial learning system might discover the association between lightning and thunder. Detecting that association is likely to be difficult if:

\begin{itemize}

\item Lightning appears in big data as a static image in one of several formats, like those mentioned above; or in a moving image in one of several formats; or it is described, in spoken or written form, as any one of such things as ``firebolt'', ``fulmination'', ``la foudre'', ``der Blitz'', ``lluched'', ``a big flash in the sky'', or indeed ``lightning''.

\item Thunder is represented in one of several different audio formats; or it is described, in spoken or written form, as ``thunder'', ``g\"ok g\"ur\"ult\"us\"u'', ``le tonnerre'', ``a great rumble'', and so on.

\end{itemize}

The association between lightning and thunder will be most easily detected via the underlying meanings of the forms that have been mentioned. We may suppose that, at some level, knowledge about lightning has an associated code or identifier, something like `\texttt{LTNG}', and that knowledge about thunder has a code or identifier such as `\texttt{THDR}'. Encodings like those would cut through much of the complexity of surface forms and allow underlying associations, such as `\texttt{LTNG THDR}', to show through.

It seems likely that at least part of the reason that people find it relatively easy to recognise, without being told, that there is an association between lightning and thunder is that, in our brains, there is some uniformity in the way different kinds of knowledge are represented and processed, without awkward inconsistencies (Section \ref{knowledge_and_brains_section}).

\subsubsection{The Interpretation of Data}\label{variety_interpretation_section}

If we are trying to recognise objects in images, do scene analysis, or otherwise interpret what the images mean, it would make things simpler if we did not have to deal with the diversity of formats for images mentioned earlier. Likewise for other kinds of data.

\subsubsection{Data Fusion}

In many fields, there is often a need to combine diverse sources of information to create a coherent whole. For example, in a study of the migration of whales, we may have, for each animal, a stream of information about the temperature of the water at each point along its route, another stream of information about the depths at which the animal is swimming, information about the weather at the surface at each point, information about dates and times, and so on.

If we are to weld those streams of information together, it would not be helpful if the geographical coordinates for different streams of information were to be expressed in different ways, perhaps using the Greenwich meridian for temperatures, the Paris meridian for depths, the Universal Transverse Mercator (UTM) system for weather, and some other scheme for the dates and times.

In short, there is a clear need to adopt a uniform system for representing the data---geographical coordinates in this example---that are needed to fuse separate but related streams of information to create a coherent view.

\subsubsection{The Understanding and Translation of Natural Languages}\label{variety_nl_section}

In our everyday use of natural languages we recognise that meanings are different from the words that express them and that, very often, two or more distinct sequences of words may mean the same thing or have the same referent: ``the capital of the United States'' means the same as ``Washington, D.~C.''; ``{\em Ursus maritimus}'' means the same as ``polar bear''; and so on. These intuitions corroborate the need for a UFK, or something like it, which is independent of the words in any natural language.

Again, it is widely recognised that, if machine translation of natural languages is ever to reach the standard of good human translators, it will be necessary to provide some kind of {\em interlingua}---an abstract language-independent representation---to express the meaning of the source language and to serve as a bridge between the source language and the target language.\footnote{\raggedright See, for example, ``Interlingual machine translation'', Wikipedia, href{http://bit.ly/1mCDTs3}{bit.ly/1mCDTs3}, retrieved 2014-01-24.} Any such interlingua is likely to be similar to or the same as a UFK.

\subsubsection{The ``Semantic Web'', the ``Internet of Things'', and the ``Web of Entities''}

The need for standardisation in the representation of knowledge is recognised in writings about the {\em semantic web} (eg, \cite{bernerslee_etal_2001}), the {\em internet of things} (eg, \cite{gershenfeld_etal_2004}), and in the Okkam project, aiming to create unique identifiers for a global {\em web of entities}.\footnote{See ``Okkam: Enabling the Web of Entities. A scalable and sustainable solution for systematic and global identifier reuse in decentralized information environments'', project reference: 215032, completed: 2010-06-30, URL: \href{http://bit.ly/OSjc1b}{bit.ly/OSjc1b}, information retrieved 2014-03-24.}

\subsubsection{The Long-Term Preservation of Data}

The continual creation of new formalisms and new formats for information and their subsequent obsolescence can mean that old data, which may include data of great value, may become unreadable or otherwise unusable. A UFK would help to reduce or eliminate this problem.

\subsubsection{Knowledge and Brains}\label{knowledge_and_brains_section}

In keeping with the long tradition in engineering of borrowing ideas from biology, the structure and functioning of brains provide reasons for trying to develop a UFK:

\begin{itemize}

\item Since brains are composed largely of neural tissue, it appears that neurons and their inter-connections, with glial cells, provide a universal framework for the representation and processing of all kinds of sensory data and all other kinds of knowledge.

\item In support of that view is evidence that one part of the brain can take over the functions of another part (see, for example, \cite{bach-y-rita_2003,bach-y-rita_kercel_2003}). This implies that there are some general principles operating across several parts of the brain, perhaps all of them.

\item Most concepts are an amalgam of several different kinds of data or knowledge. For example, the concept of a ``picnic'' combines the sights, sounds, tactile and gustatory sensations, and the social and logistical knowledge associated with such things as a light meal in pleasant rural surroundings. To achieve that kind of seamless integration of different kinds of knowledge, it seems necessary for the human brain to be or to contain a UFK.

\end{itemize}

\subsection{The Potential of the SP System as a Universal Framework for the Representation and Processing of Knowledge}\label{sp_as_ufk_section}

In the SP programme, the aim has been to create a system that, in accordance with Occam's Razor, combines conceptual {\em simplicity} with descriptive or explanatory {\em power} \cite[Section 1.3]{wolff_2006}, \cite[Section 2]{sp_benefits_apps}. Although the SP computer model is relatively simple---its ``exec'' file requires less than 500 KB of storage space---and despite the great simplicity of SP {\em patterns} as a vehicle for knowledge (Section \ref{introduction_to_sp_section}), the SP system, without additional programming, may serve in the representation and processing of several different kinds of knowledge:

\begin{itemize}

\item {\em Syntax and semantics of natural languages}. The system provides for the representation of syntactic rules, including discontinuous dependencies in syntax, and for the parsing and production of language \cite[Chapter 5]{wolff_2006}, \cite[Section 8]{sp_extended_overview}. It has potential to represent non-syntactic `meanings' via such things as class hierarchies and part-whole hierarchies (next), and it has potential in the understanding of natural language and in the production of sentences from meanings \cite[Section 5.7]{wolff_2006}.

\item {\em Class hierarchies and part-whole hierarchies}. The system lends itself to the representation of class hierarchies (eg, species, genus, family, etc), heterarchies (class hierarchies with cross-classification), and part-whole hierarchies (eg, [[head [eyes, nose, mouth, ...]], [body ...], [legs ...]]) and their processing in pattern recognition, reasoning, and more \cite[Sections 6.4.1 and 6.4.2]{wolff_2006}, \cite[Section 9.1]{sp_extended_overview}.

\item {\em Networks and trees}. The SP system supports the representation and processing of such things as hierarchical and network models for databases \cite[Section 5]{wolff_sp_intelligent_database}, and probabilistic decision networks and decision trees \cite[Section 7.5]{wolff_2006}. And it has advantages as an alternative to Bayesian networks \cite[Section 7.8]{wolff_2006} (reproduced in \cite[Sections 10.2, 10.3, and 10.4]{sp_extended_overview}).

\item {\em Relational knowledge}. The system supports the representation of knowledge with relational tuples, and retrieval of information in the manner of query-by-example \cite[Section 3]{wolff_sp_intelligent_database}, and it has some apparent advantages compared with the relational model \cite[Section 4.2.3]{wolff_sp_intelligent_database}.

\item {\em Rules and reasoning}. The system supports several kinds of reasoning, with the representation of associated knowledge. Examples include one-step `deductive' reasoning, abductive reasoning, chains of reasoning, reasoning with rules, nonmonotonic reasoning, and causal diagnosis \cite[Chapter 7]{wolff_2006}.

\item {\em Patterns and pattern recognition}. The SP system has strengths in the representation and processing of one-dimensional patterns \cite[Chapter 6]{wolff_2006}, \cite[Section 9]{sp_benefits_apps}, and it may be applied to medical diagnosis, viewed as a pattern recognition problem \cite{wolff_medical_diagnosis}.

\item {\em Images}. Although the SP computer model has not yet been generalised to work with patterns in two dimensions, there is clear potential for the SP system to be applied to the representation and processing of images and other kinds of information with a 2D form. This is discussed in \cite[Section 13.2.1]{wolff_2006} and also in \cite{sp_vision}.

\item {\em Structures in three dimensions}. It appears that the multiple alignment framework may be applied to the representation and processing of 3D structures via the stitching together of overlapping 2D views \cite[Section 7.1]{sp_vision}, in much the same way that 3D models may be created from overlapping 2D photos,\footnote{\raggedright See, for example, ``Big Object Base'' (\href{http://bit.ly/1gwuIfa}{bit.ly/1gwuIfa}), ``Camera 3D'' (\href{http://bit.ly/1iSEqZu}{bit.ly/1iSEqZu}), or ``PhotoModeler'' (\href{http://bit.ly/MDj70X}{bit.ly/MDj70X}.)} or a panoramic photo may be created from overlapping shots.

\item {\em Procedural knowledge}. The SP system can represent simple procedures (actions that need to be performed in a particular sequence); it can model such things as `variables', `values', `types', `function with parameters', repetition of operations, and more \cite[Section 6.6.1]{sp_benefits_apps}; and it has potential to represent sets of procedures that may be performed in parallel \cite[Section 6.6.3]{sp_benefits_apps}. These representations may serve to control real-world operations in sequence and in parallel.

\end{itemize}

As a candidate for the role of UFK, the SP system has other strengths:

\begin{itemize}

\item Because of the generality of the concept of information compression via the matching and unification of patterns, there is reason to believe that the system may be applied to the representation and processing of all kinds of knowledge, not just those listed above.

\item Because all kinds of knowledge are represented in one simple format (arrays of atomic symbols in one or two dimensions), and because all kinds of knowledge are processed in the same way (via the creation of multiple alignments), the system provides for the seamless integration of diverse kinds of knowledge, in any combination \cite[Section 7]{sp_benefits_apps}.

\item Because of the system's existing and potential capabilities in learning and discovery (Section \ref{learning_discovery_section}), it has potential for the automatic structuring of knowledge, reducing or eliminating the need for hand crafting, with corresponding benefits in terms of speed, cost, and reducing errors.

\item For reasons given in Section \ref{visualisation_section}), the SP system may facilitate the visualisation of structures and processes via static and moving images.

\end{itemize}

In summary, the relative simplicity of the SP system, its versatility in the representation and processing of diverse kinds of knowledge, its provision for seamless integration of different kinds of knowledge in any combination, the system's potential for automatic structuring of knowledge, and for the visualisation of structures and processes, makes it a good candidate for development into a UFK.

\subsection{Standardisation and Translation}\label{standardisation_translation_section}

The SP system, or any other UFK, may be used in two distinct ways:

\begin{itemize}

\item {\em Standardisation in the representation of knowledge}. There is potential, on relatively long timescales, to standardise the representation and processing of many kinds of knowledge, cutting out much of the current jumble of formalisms and formats. But for the kinds of reasons mentioned in Section \ref{problem_of_variety_section}, it is likely that some of those formalisms or formats will never be replaced or will co-exist with representation and processing via the UFK.

\item {\em Translation into the universal framework}. Where a body of information is expressed in one or more non-standard forms but is needed in the standard form, it may be translated. This may be done via the SP system, as outlined in Section \ref{interpretation_section}. Or it may be done using conventional technologies, in much the same way that the source code for a computer program may, using a compiler, be translated into object code. The translation of natural languages is likely to prove more challenging than the translation of artificial formalisms and formats.

\end{itemize}

Either way, any body of big data may be expressed in a standard form that facilitates the unsupervised learning or discovery of structures and associations within those data (Section \ref{learning_discovery_section}), and facilitates forms of interpretation as outlined in Section \ref{interpretation_section}.

\section{Learning and Discovery}\label{learning_discovery_section}

\begin{quote}

``While traditional computers must be programmed by humans to perform specific tasks, cognitive systems will learn from their interactions with data and humans and be able to, in a sense, program themselves to perform new tasks.'' \cite[p.~7]{kelly_hamm_2013}.

\end{quote}

In broad terms, unsupervised learning in the SP system means lossless compression of a body of information, {\bf I}, via the matching and unification of patterns (Section \ref{product_of_learning_section}).

The SP computer model (SP70, \cite[Chapter 9]{wolff_2006}, \cite[Section 5]{sp_extended_overview}), has already demonstrated an ability to discover generative grammars from unsegmented samples of English-like artificial languages, including segmental structures, classes of structure, and abstract patterns. As it is now, it has shortcomings, outlined in \cite[Section 3.3]{sp_extended_overview}. But I believe these problems are soluble, and that their solution will clear the path to the unsupervised learning of other kinds of structures, such as class hierarchies, part-whole hierarchies, and discontinuous dependencies in data. In what follows, we shall assume that these and other problems have been solved and that the system is relatively robust and mature.

A strength of the SP system is that it can discover structures in data, not merely statistical associations between pre-established structures.

As noted in Section \ref{introduction_to_sp_section}, evidence to date suggests that the system conforms to the DONSVIC principle \cite[Section 5.2]{sp_extended_overview}.

\subsection{The Product of Learning}\label{product_of_learning_section}

The product of learning from a body of data, {\bf I}, may be seen to comprise a {\em grammar} ({\bf G}) and an {\em encoding} ({\bf E}) of {\bf I} in terms of the grammar. Here, the term `grammar' has a broad meaning that includes grammars for natural languages, grammars for static and moving images, grammars for business procedures, and so on.

As with all other kinds of data in the SP system, {\bf G} and {\bf E} are both represented using SP patterns.

In accordance with the principles of {\em minimum length encoding} \cite{solomonoff_1964}, the SP system aims to minimise the overall size of {\bf G} and {\bf E}.\footnote{\raggedright The similarity with research on grammatical inference is not accidental: the SP programme of research has grown out of earlier research developing computer models of language learning (see \cite{wolff_1988} and other publications that may be downloaded via \href{http://bit.ly/JCd6jm}{bit.ly/JCd6jm}). But in developing the SP system, a radical reorganisation has been needed to meet the goal of simplifying and integrating concepts across artificial intelligence, mainstream computing, and human perception and cognition. Unlike the earlier models and other research on grammatical inference, multiple alignment is central in the workings of the SP computer model, including unsupervised learning. A bonus of the new structure is potential for the unsupervised learning of such things as class hierarchies, part-whole hierarchies, and discontinuous dependencies in data.} Together, {\bf G} and {\bf E} achieve lossless compression of {\bf I}.

{\bf G} is largely about {\em redundancies} within {\bf I}, while {\bf E} is mainly a record of the non-redundant aspects of {\bf I}. Here, any symbol or sequence of symbols represents redundancy within {\bf I} if it repeats more often than one would expect by chance. To reach that threshold, small patterns need to occur more frequently than large patterns.\footnote{\raggedright {\bf G} may contain some patterns that do not, in themselves, represent redundancy but are included in {\bf G} because of their supporting role \cite[Section 3.6.2]{wolff_2006}.}

{\bf G} may be regarded as a distillation of the `essence' of {\bf I}. Normally, {\bf G} would be more interesting than {\bf E}, and more useful in the kinds of applications described in Sections \ref{interpretation_section} and \ref{learning_with_errors_uncertainties_section}.

With data that is received or processed as a stream rather than a static body of data of fixed size (Section \ref{velocity_section}, below), {\bf G} may be grown incrementally. And, quite often, there is likely to be a case for merging {\bf G}s from different sources, with unification of patterns that are the same. In principle, there could be a single `super' {\bf G}, expressing the essentials of the world's knowledge in a compressed form. Similar remarks apply to {\bf E}s---if they are needed.

\subsection{Unsupervised Learning and the Problem of Variety in Big Data}\label{unsupervised_learning_and_variety_section}

Systems for unsupervised learning may be applied most simply and directly when the data for learning come in a uniform style as, for example, in DNA data: simple sequences of the letters \texttt{A}, \texttt{T}, \texttt{G}, and \texttt{C}. But as outlined in Section \ref{variety_discovery_section}, it may be difficult to discover recurrent associations or structures when there is a variety of formalisms and formats.

The discussion here focuses on the relatively challenging area of natural languages, because the variety of natural languages is a significant part of the problem of variety in big data, because the SP system has strengths in that area, and because it seems likely that solutions with natural languages will generalise relatively easily to other areas.

With natural languages, learning processes in a mature version of the SP system may be applied in four distinct but inter-related ways, discussed in the following subsections.

\subsubsection{Learning the Surface Forms of Language}\label{surface_forms_section}

If the data for learning are text in a natural language, then the product of learning (Section \ref{product_of_learning_section}) will be about words and parts of words, about phrases, clauses and sentences, and about grammatical categories at all levels. Likewise with speech.

Even with human-like capabilities in learning, a {\bf G} that is derived without the benefit of meanings is likely to differ in some respects from a grammar created by a linguist who can understand what the text means. This is because there are subtle interdependencies between syntax and semantics \cite[Section 6.2]{sp_benefits_apps} which cannot be captured from text on its own, without information about meanings.

\subsubsection{Learning Non-Syntactic Knowledge}\label{non-syntactic_section}

The SP system may be applied to learning about the non-syntactic world: objects and their interactions, scenery, music, games, and so on. These have an intrinsic interest that is independent of natural language, but they are also the things that people talk and write about: the non-syntactic meanings or semantics of natural language. Some aspects of this area of learning are discussed in \cite[Section 13.2.1]{wolff_2006} and \cite{sp_vision}.

\subsubsection{Connecting Syntax with Semantics}

Of course, for any natural language to be effective, syntax must connect with semantics. Examples that show how syntax and semantics may work together in the multiple alignment framework, in both the analysis and production of language, are presented in \cite[Section 5.7]{wolff_2006}. As noted in Section \ref{sp_as_ufk_section}, seamless integration of different kinds of knowledge is facilitated by the use of one simple format for all kinds of knowledge and a single framework---multiple alignment---for processing diverse kinds of knowledge.

In broad terms, making the connection between syntax and semantics means associational learning, no different in principle from learning the association between lightning and thunder (Section \ref{variety_discovery_section}), between smoke and fire, between a savoury aroma and a delicious meal, and so on.

For the SP system to learn the connections between syntax and semantics, it will need speech or text to be presented alongside the non-syntactic information that it relates to, in much the same way that, normally, children have many opportunities to hear people talking and see what they are talking about at the same time.

\subsubsection{Learning Via the Interpretation of Surface Forms}

Since speech and text in natural languages are an important part of big data, it is clear that if the SP system, or any other learning system, is to get full value from big data, it will need to learn from the meanings of speech or text as well as from their surface forms.

For any given body of text or speech, {\bf I}, the first step, of course, will be to derive its meanings. This can be done via processes of interpretation, as described in Section \ref{interpretation_section}.

The set of SP patterns that represent the meanings of {\bf I} may then be processed as if it was New information, searching for redundancies in the data, unifying patterns that match each other, and creating a compressed representation of the data. Then it should be possible to discover such things as the association between lightning and thunder (Section \ref{variety_discovery_section}), regardless of how the data was originally presented.

\section{Interpretation of Data}\label{interpretation_section}

By contrast with unsupervised learning, which compresses a body of information ({\bf I}) to create {\bf G} and {\bf E}, the concept of {\em interpretation} in this article means processing {\bf I} in conjunction with a pre-established grammar ({\bf G}) to create a relatively compact encoding ({\bf E}) of {\bf I}.

Depending on the nature of {\bf I} and {\bf G}, the process of interpretation may be seen to achieve such things as pattern recognition, information retrieval, parsing or production of natural language, translation from one representation to another, several kinds of reasoning, planning, and problem solving. Some of these were touched on briefly in Section \ref{sp_as_ufk_section}. Here is a little more detail:

\begin{itemize}

\item {\em Pattern recognition}. With the SP system, pattern recognition may be achieved: at multiple levels of abstraction; with ``family resemblance'' or polythetic categories; in the face of errors of omission, commission or substitution in data; with the calculation of a probability for any given identification, classification or associated inference; with sensitivity to context in recognition; and with the seamless integration of pattern recognition with other aspects of intelligence---reasoning, learning, problem solving, and so on \cite[Section 9]{sp_extended_overview}, \cite[Chapter 6]{wolff_2006}. As previously mentioned, the system may be applied in computer vision \cite{sp_vision} and in medical diagnosis \cite{wolff_medical_diagnosis}, viewed as pattern recognition.

\item {\em Information retrieval}. The SP system lends itself to information retrieval in the manner of query-by-example and, with the provision of SP patterns representing relevant rules, there is potential to create the facilities of a query language like SQL \cite{wolff_sp_intelligent_database}.

\item {\em Parsing and production of natural language}. As can be seen in Fig.~\ref{parsing_1_figure}, the creation of a multiple alignment in the SP system may achieve the effect of parsing a sentence in a natural language (see also \cite[Section 3.4.3 and Chapter 5]{wolff_2006}). It may also function in the production of sentences \cite[Section 3.8]{wolff_2006}.

\item {\em Translation from one representation to another}. There is potential with the SP system for the integration of syntax and semantics in both the understanding and production of natural language \cite[Section 5.7]{wolff_2006}, with corresponding potential for the translation of any one language into an SP-style interlingua and further translation into any other natural language \cite[Section 6.2.1]{sp_benefits_apps}. Probably less challenging, as mentioned earlier, would be the translation of artificial formalisms and formats---JPEG, MP3, and so on---into an SP-style representation.

\item {\em Several kinds of reasoning}. The SP system can perform several kinds of reasoning, including one-step `deductive' reasoning, abductive reasoning, reasoning with probabilistic networks and trees, reasoning with `rules', nonmonotonic reasoning, Bayesian reasoning and ``explaining away'', causal diagnosis, and reasoning that is not supported by evidence \cite[Chapter 7]{wolff_2006}.

\item {\em Planning}. With SP patterns representing direct flights between cities, the SP system can normally work out one or more routes between any two cities that are not connected directly, if such a route exists \cite[Section 8.2]{wolff_2006}.

\item {\em Problem solving}. The system can also solve textual versions of geometric analogy problems, like those found in puzzle books and IQ tests \cite[Section 8.3]{wolff_2006}.

\end{itemize}

\section{Velocity: Analysis of Streaming Data}\label{velocity_section}

\begin{quote}

``Most of today's computing tasks involve data that have been gathered and stored in databases. The data make a stationary target. But, increasingly, vitally important insights can be gained from analyzing information that's on the move.~...~This approach is called streams analytics. Rather than placing data in a database first, the computer analyses it as it comes from a variety of sources, continually refining its understanding of the data as conditions change. This is the way humans process information.'' \cite[pp.~49--50]{kelly_hamm_2013}.

\end{quote}

Although, in its unsupervised learning, the SP system may process information in batches, it lends itself most naturally to an incremental style. In the spirit of the quotation above, the SP system is designed to assimilate {\em New} information to a steadily-growing body of relatively-compressed {\em Old} information, as shown schematically in Fig.~\ref{sp_input_perspective_figure}.

Likewise, in interpretive processes such as pattern recognition, processing of natural language, and reasoning, the SP system may be applied to streams of data as well as the processing of data in batches.

\section{Volume: Making Big Data Smaller}\label{making_big_data_smaller_section}

\begin{quote}

``Very-large-scale data sets introduce many data management challenges.'' \cite[p.~41]{national_research_council_2013}.

\end{quote}

\begin{quote}

``In addition to reducing computation time, proper data representations can also reduce the amount of required storage (which translates into reduced communication if the data are transmitted over a network).'' \cite[p.~68]{national_research_council_2013}.

\end{quote}

Because information compression is central in how the SP system works, it has potential to reduce problems of volume in big data by making it smaller. Although comparative studies have not yet been attempted, the SP system has potential to achieve relatively high levels of lossless compression for two main reasons: it is designed so that, if required, it can perform a relatively thorough search for redundancies in data; and there is potential to tap into discontinuous dependencies in data, an aspect of redundancy that appears to be outside the scope of other systems for compression of information \cite[Section 6.7]{sp_benefits_apps}.

In brief, information compression in the SP system can yield {\em direct benefits} in the storage, management, and transmission of data, and {\em indirect benefits} as described elsewhere in this article: unsupervised learning (Section \ref{learning_discovery_section}), processes of interpretation such as pattern recognition and reasoning (Section \ref{interpretation_section}), economies in the transmission of data via the separation of grammar and encoding (Section \ref{transmission_section}), gains in computational efficiency (Section \ref{computational_efficiency_section}), and assistance in the management of errors and uncertainties in data (Section \ref{errors_uncertainties_section}).

\section{Additional Economies in the Transmission of Data}\label{transmission_section}

\begin{quote}

``One roadblock to using cloud services for massive data analysis is the problem of transferring the large data sets. Maintaining a high-capacity and wide-scale communications network is very expensive and only marginally profitable.'' \cite[p.~55]{national_research_council_2013}.

\end{quote}

\begin{quote}

``To control costs, designers of the [DOME] computing system have to figure out how to minimize the amount of energy used for processing data. At the same time, since so much of the energy in computing is required to move data around, they have to discover ways to move the data as little as possible.'' \cite[p.~65]{kelly_hamm_2013}.

\end{quote}

Although the second of these two quotes may refer in part to movements of data such as those between the CPU and the memory of a computer, the discussion here is about transmission of data over longer distances such as, for example, via the internet.

As we have seen (Section \ref{making_big_data_smaller_section}), the SP system may promote the efficient transmission of data by making it smaller. But there is potential with the SP system for additional economies in the transmission of data and these may be very substantial \cite[Section 6.7.1]{sp_benefits_apps}.

Any body of data, {\bf I}, may be compressed by encoding it in terms a `grammar' ({\bf G}), provided that {\bf G} contains the kinds of structures that are found in {\bf I} (Section \ref{product_of_learning_section}). Then {\bf I} may be sent from A to B by sending only the `encoding' ({\bf E}). Provided that B has a copy of {\bf G}, {\bf I} may be recreated with complete fidelity by means of the SP system \cite[Section 4.5]{sp_extended_overview}, \cite[Section 3.8]{wolff_2006}. Since {\bf E} would normally be very much smaller than the {\bf I} from which it was derived, it seems likely that there would be a net gain in efficiency, allowing for the computational costs of encoding and decoding.

Since a copy of {\bf G} must be transmitted to B, any savings will be relatively small if it is used only for the decoding of a single instance of {\bf E}. But significant savings are likely if, as would normally be the case, one copy of {\bf G} may be used for the decoding of many different instances of {\bf E}, representing many different {\bf I}s.

In this kind of application, it would probably make sense for there to be a division of labour between creating a grammar and using it in the encoding and decoding of data. For example, the computational heavy lifting required to build a grammar for video images may be done by a high-performance computer. But that grammar, once constructed, may serve in relatively low-powered devices---smartphones, tablet computers, and the like---for the much less demanding processes of encoding and decoding video transmissions.

\section{Energy, Speed, and Bulk}\label{computational_efficiency_section}

\begin{quote}

``... we're reaching the limits of our ability to make [gains in the capabilities of CPUs] at a time when we need even more computing power to deal with complexity and big data. And that's putting unbearable demands on today's computing technologies---mainly because today's computers require so much energy to perform their work.'' \cite[p.~9]{kelly_hamm_2013}.

\end{quote}

\begin{quote}

``The human brain is a marvel. A mere 20 watts of energy are required to power the 22 billion neurons in a brain that's roughly the size of a grapefruit. To field a conventional computer with comparable cognitive capacity would require gigawatts of electricity and a machine the size of a football field. So, clearly, something has to change fundamentally in computing for sensing machines to help us make use of the millions of hours of video, billions of photographs, and countless sensory signals that surround us. ... Unless we can make computers many orders of magnitude more energy efficient, we're not going to be able to use them extensively as our intelligent assistants.'' \cite[p.~75, p.~88]{kelly_hamm_2013}.

\end{quote}

\begin{quote}

``Supercomputers are notorious for consuming a significant amount of electricity, but a less-known fact is that supercomputers are also extremely `thirsty' and consume a huge amount of water to cool down servers through cooling towers ....''\footnote{\raggedright From ``How can supercomputers survive a drought?'', HPC Wire, 2014-01-26, \href{http://bit.ly/LruEPS}{bit.ly/LruEPS}.}

\end{quote}

It is now clear that, if we are to do meaningful analyses of more than a small fraction of present and future floods of big data, substantial gains will be needed in the computational efficiency of computers, with associated benefits:

\begin{itemize}

\item Cutting demands for energy, with corresponding cuts in the need for cooling of computers.

\item Speeding up processing with a given computational resource.

\item Consequent reductions in the size and weight of computers.

\end{itemize}

With the possible exception of the need for cooling, these things are particularly relevant to mobile devices, including autonomous robots.

The following subsections describe how the SP system may contribute to computational efficiency, via information compression and probabilities, and via a synergy with `data-centric' computing.

\subsection{Computational Efficiency via Information Compression and Probabilities}\label{efficiency_ic_section}

In the light of evidence that the SP system is Turing-equivalent \cite[Chapter 4]{wolff_2006}, and since information processing in the SP system means compression of information via the matching and unification of patterns (Section \ref{introduction_to_sp_section}), {\em anything that increases the efficiency of searching for good full and partial matches between patterns will also increase the efficiency of information processing}.

It appears that information compression and associated probabilities can themselves be a means of increasing the efficiency of searching, as described in the next two subsections.

\subsubsection{Reducing the Sizes of Data to be Searched and of Search Terms}\label{reducing_data_sizes_section}

As described in \cite[Section 6.7.2]{sp_benefits_apps}, if we wish to search a body of information, {\bf I}, for instances of a pattern like ``Treaty on the Functioning of the European Union,'' the efficiency of searching may be increased:

\begin{itemize}

\item By reducing the size of {\bf I} so that there is less to be searched. The size of {\bf I} may be reduced by replacing all but one of the instances of ``Treaty on the Functioning of the European Union'' with a relatively short code or identifier like ``TFEU'', and likewise with other recurrent patterns. More generally, the size of {\bf I} may be reduced via unsupervised learning in the SP system. It is true that the compression of {\bf I} is a computational cost, but this investment is likely to pay off in later processing.

\item By searching with a short code like ``TFEU'' instead of a relatively large pattern like ``Treaty on the Functioning of the European Union''. Other things being equal, a smaller search pattern means faster searching.

\end{itemize}

With regard to the second point, there is potential to cut out some searching altogether by creating direct connections between each instance of a code (``TFEU'' in this example) and the thing that it represents (``Treaty on the Functioning of the European Union''). In {\em SP-neural} (Section \ref{sp_neural_section}), there are connections of that kind between ``pattern assemblies'', as shown schematically in Fig.~\ref{class_part_figure}.

\subsubsection{Concentrating Search Where Good Results Are Most Likely to be Found}\label{concentrating_search_section}

If we want to find some strawberry jam, our search is more likely to be successful in a supermarket than it would be in a hardware shop or a car-sales showroom. This may seem too simple and obvious to deserve comment but it illustrates the extraordinary knowledge that most people have of an informal `statistics' of the world that we inhabit, and how that knowledge may help us to minimise effort.\footnote{\raggedright See also G.~K.~Zipf's {\em Human Behaviour and the Principle of Least Effort} \cite{zipf_1949}.}

Where does that statistical knowledge come from? In the SP theory, it flows directly from the central role of information compression in our perceptions, learning and thinking, and from the intimate relationship between information compression and concepts of prediction and probability \cite{li_vitanyi_2009}.

Although the SP computer model may calculate probabilities associated with multiple alignments (Section \ref{introduction_to_sp_section}), it actually uses levels of information compression as a guide to search. Those levels are used, with heuristic search methods (including escape from `local peaks'), to ensure that searching is concentrated in areas where it is most likely to be fruitful \cite[Sections 3.9, 3.10, and 9.2]{wolff_2006}. This not only speeds up processing but yields Big-O values for computational complexity that are within acceptable limits \cite[Sections 3.10.6, 9.3.1, and A.4]{wolff_2006}.

\subsubsection{Potential Gains in Computational Efficiency}\label{potential_efficiency_ic_section}

No attempt has yet been made to quantify potential gains in computational efficiency from the compression of information, as described in Sections \ref{reducing_data_sizes_section}, \ref{concentrating_search_section}, and \ref{transmission_section}, but they could be very substantial:

\begin{itemize}

\item Since information compression is fundamental in the workings of the SP system, there is potential for corresponding savings in all parts and levels in the system.

\item The entire structure of knowledge that the system creates for itself is intrinsically statistical, with potential on many fronts for corresponding savings in computational costs and associated demands for energy.

\end{itemize}

It may be argued that, since object-oriented programming already provides for compression of information via class hierarchies and inheritance of attributes, the benefits of information compression are already available in conventional computing systems. In response, it may be said that, while there are undoubted benefits from object-oriented programming, existing object-oriented systems run on conventional computers and suffer from the associated inefficiencies.

Realising the full potential of information compression as a means of improving computational efficiency will probably mean new thinking about computer architectures, probably in conjunction with the development of data-centric computing (next).

\subsection{A Potential Synergy with Data-Centric Computing}\label{data-centric_computing_section}

\begin{quote}

``What's needed is a new architecture for computing, one that takes more inspiration from the human brain. Data processing should be distributed throughout the computing system rather than concentrated in a CPU. The processing and the memory should be closely integrated to reduce the shuttling of data and instructions back and forth.'' \cite[p.~9]{kelly_hamm_2013}.

\end{quote}

\begin{quote}

``Unless we can make computers many orders of magnitude more energy efficient, we're not going to be able to use them extensively as our intelligent assistants. Computing intelligence will be too costly to be practical. Scientists at IBM Research believe that to make computing sustainable in the era of big data, we will need a different kind of machine---the data-centric computer. ... Machines will perform computations faster, make sense of large amounts of data, and be more energy efficient.'' \cite[p.~88]{kelly_hamm_2013}.

\end{quote}

The SP concepts may help to integrate processing and memory, as described in the next two subsections.

\subsubsection{SP-Neural}\label{sp_neural_section}

Although the main emphasis in the SP programme has been on developing an abstract framework for the representation and processing of knowledge, the theory includes proposals---called {\em SP-neural}---for how those abstract concepts may be realised with neurons \cite[Chapter 11]{wolff_2006}.

\begin{figure*}[!hbt]
\centering
\includegraphics[width=0.6\textwidth]{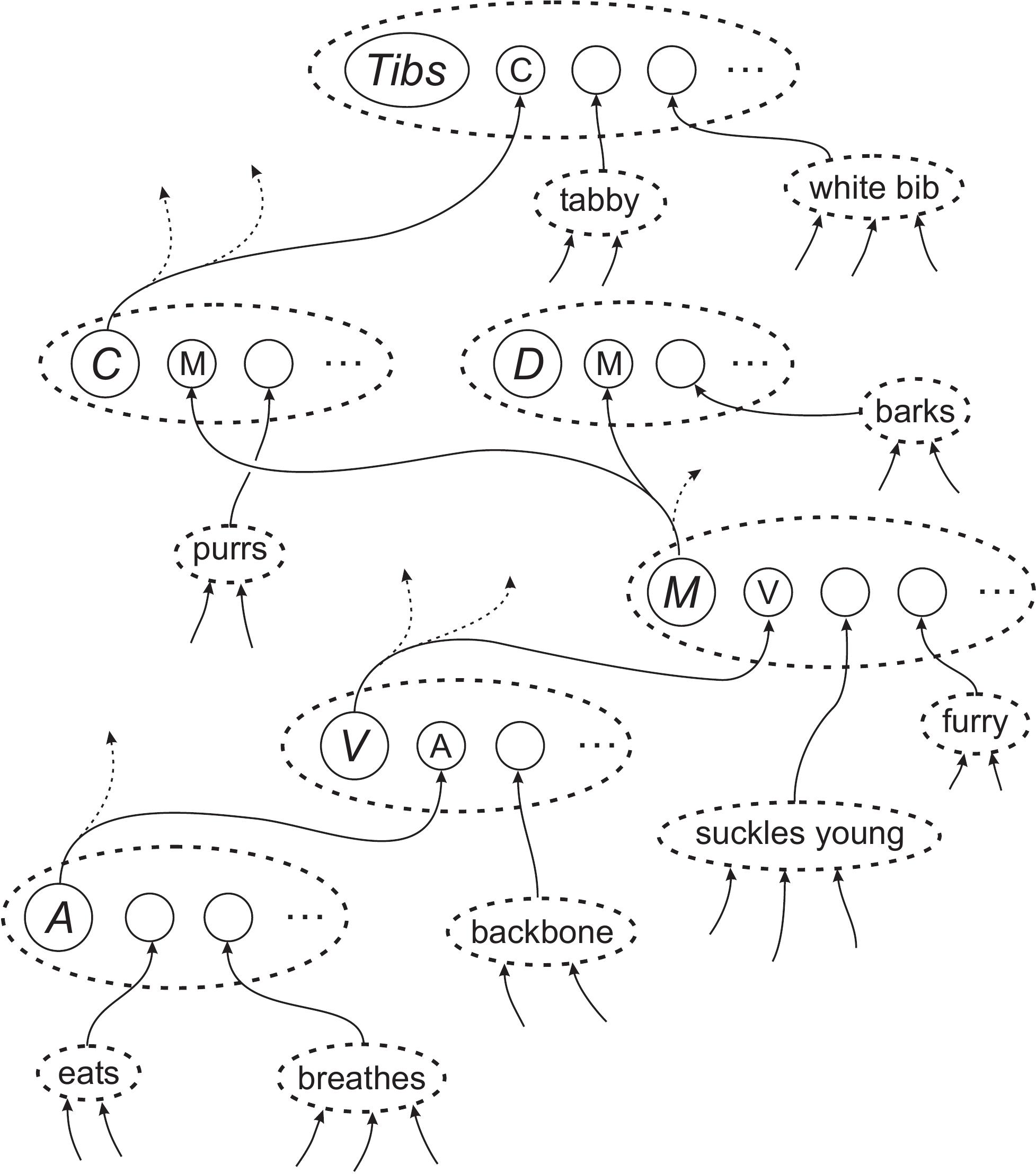}
\caption{An example showing schematically how SP-neural may represent class-inclusion relations, part-whole relations, and their integration. {\em Key}: `C' = cat, `D' = dog, `M' = mammal, `V' = vertebrate, `A' = animal, `...' = further structure that would be shown in a more comprehensive example. Pattern assemblies are surrounded by broken lines and each neuron is represented by an unbroken circle or ellipse. Lines with arrows show connections between pattern assemblies and the flow of sensory signals in the process of recognising something (there may also be connections in the opposite direction to support the production of patterns). Connections between neurons within each pattern assembly are not marked. Reproduced from Fig.~11.6 in \cite{wolff_2006}, with permission.}
\label{class_part_figure}
\end{figure*}

Fig.~\ref{class_part_figure} shows in outline how an SP-style conceptual structure would appear in SP-neural. It is envisaged that SP patterns would be realised with {\em pattern assemblies}---groupings of neurons like those shown in the figure within broken-line envelopes.

The whole scheme is quite different from `artificial neural networks' as they are commonly conceived in computer science.\footnote{\raggedright See, for example, ``Artificial neural network'', Wikipedia, \href{http://en.wikipedia.org/wiki/Artificial\_neural\_network}{en.wikipedia.org/wiki/Artificial\_neural\_network}, retrieved 2013-12-23.} It may be seen as a development of Donald Hebb's \cite{hebb_1949} concept of a `cell assembly', with more precision about how structures may be shared, and other differences.\footnote{\raggedright In particular, unsupervised learning in the SP system \cite[Section 5]{sp_extended_overview}, \cite[Chapter 9]{wolff_2006} is radically different from the ``Hebbian'' concept of learning (see, for example, ``Hebbian theory'', Wikipedia, \href{en.wikipedia.org/wiki/Hebbian\_learning}{http://en.wikipedia.org/wiki/Hebbian\_learning}, retrieved 2013-12-23), described by Hebb \cite{hebb_1949} and adopted as the mechanism for learning in most artificial neural networks. By contrast with Hebbian learning, the SP system, like a person, may learn from a single exposure to some situation or event. And, by contrast with Hebbian learning, it takes time to learn a language in the SP system because of the complexity of the search space, not because of any kind of gradual strengthening or ``weighting'' of links between neurons \cite[Section 11.4.4]{wolff_2006}.}

In SP-neural, what is essentially a statistical model of the world is reflected directly in groupings of neurons and their interconnections, as shown in Fig.~\ref{class_part_figure}. It is envisaged that such things as pattern recognition would be achieved via the transmission of impulses between pattern assemblies, and via the transmission of impulses between neurons within each pattern assembly. In keeping with what is known about the workings of brains and nervous systems, it is likely that there would be important roles for both excitatory and inhibitory signals.

In short, neurons in SP-neural serve for both the representation and processing of knowledge, with close integration of the two---in accordance with the concept of data-centric computing. One architecture may promote computational efficiency by combining the benefits of information compression and probabilistic knowledge with the benefits of data-centric computing.

\subsubsection{Computing with Light or Chemicals}

The SP concepts appear to lend themselves to computing with light or chemicals, perhaps by-passing such things as transistors or logic gates that have been prominent in the development of electronic computers \cite[Section 6.10.6]{sp_benefits_apps}.\footnote{\raggedright ``The most promising means of moving data faster is by harnessing photonics, the generation, transmission, and processing of light waves.'' \cite[p.~93]{kelly_hamm_2013}.}

At the heart of the SP system is a process of finding good full and partial matches between patterns. This may be done with light, with the potential advantage that light beams may cross each other without interference. Another potential advantage is that, with collimated light, there may be relatively small losses over distance---although distances should probably be minimised to save on transmission times and to minimise the sizes of computing devices. There appears to be potential to create an optical or optical/electronic version of SP-neural.

Finding good full and partial matches between patterns may also, potentially, be done with chemicals such as DNA,\footnote{\raggedright See, for example, ``DNA computing'', Wikipedia, \href{http://bit.ly/1gfEP4p}{bit.ly/1gfEP4p}, retrieved 2013-12-30.} with potential for high levels of parallelism, and with the attraction that DNA can be a means of storing information in a very compact form, and for very long periods \cite{goldman_etal_2013}.

With both light and chemicals, the SP system may help realise data-centric integration of knowledge and processing. As before, there is potential for gains in computational efficiency via one architecture that combines the benefits of information compression and probabilistic knowledge with the benefits of data-centric computing.

\section{Veracity: Managing Errors and Uncertainties in Data}\label{errors_uncertainties_section}

\begin{quote}

``In building a statistical model from any data source, one must often deal with the fact that data are imperfect. Real-world data are corrupted with noise. Such noise can be either systematic (i.e., having a bias) or random (stochastic). Measurement processes are inherently noisy, data can be recorded with error, and parts of the data may be missing.'' \cite[p.~99]{national_research_council_2013}.

\end{quote}

\begin{quote}

``Organizations face huge challenges as they attempt to get their arms around the complex interactions between natural and human-made systems. The enemy is uncertainty. In the past, since computing systems didn't handle uncertainty well, the tendency was to pretend that it didn't exist. Today, it is clear that that approach won't work anymore. So rather than trying to eliminate uncertainty, people have to embrace it.'' \cite[pp.~50--51]{kelly_hamm_2013}.

\end{quote}

The SP system has potential in the management of errors and uncertainties in data as described in the following subsections.

\subsection{Parsing or Pattern Recognition That Is Robust in the Face of Errors}\label{robust_against_errors_section}

As mentioned in Section \ref{introduction_to_sp_section}, the SP system is inherently probabilistic. Every SP pattern has an associated frequency of occurrence, and probabilities may be derived for each multiple alignment \cite[Section 4.4]{sp_extended_overview}, \cite[Section 3.7 and Chapter 7]{wolff_2006}.

The probabilistic nature of the system means that, in operations such as parsing natural language or pattern recognition, it is robust in the face of errors of omission, of commission, or of substitution \cite[Section 4.2.2]{sp_extended_overview}, \cite[Section 6.2.1]{wolff_2006}. In the same way that we can recognise things visually despite disturbances such as falling leaves or snow (and likewise for other senses), the SP system can, within limits, produce what we intuitively judge to be `correct' analyses of inputs that are not entirely accurate.

\begin{figure*}[!htbp]
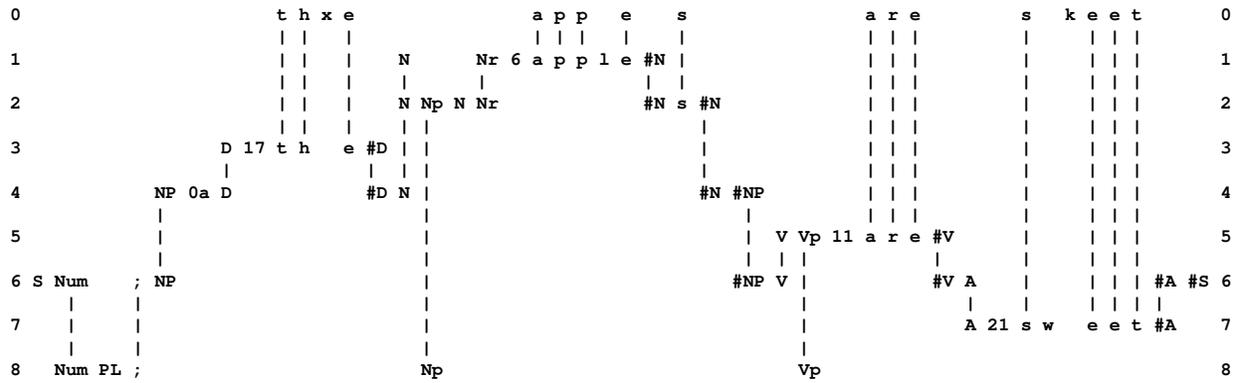

\fontsize{07.00pt}{08.40pt}
\centering
{\bf
\begin{BVerbatim}
0                       t h x e                a p p   e    s                a r e         s   k e e t       0
                        | |   |                | | |   |    |                | | |         |     | | |
1                       | |   |    N      Nr 6 a p p l e #N |                | | |         |     | | |       1
                        | |   |    |      |              |  |                | | |         |     | | |
2                       | |   |    N Np N Nr             #N s #N             | | |         |     | | |       2
                        | |   |    | |                        |              | | |         |     | | |
3                  D 17 t h   e #D | |                        |              | | |         |     | | |       3
                   |            |  | |                        |              | | |         |     | | |
4            NP 0a D            #D N |                        #N #NP         | | |         |     | | |       4
             |                       |                            |          | | |         |     | | |
5            |                       |                            |  V Vp 11 a r e #V      |     | | |       5
             |                       |                            |  | |           |       |     | | |
6 S Num    ; NP                      |                           #NP V |           #V A    |     | | | #A #S 6
     |     |                         |                                 |              |    |     | | | |
7    |     |                         |                                 |              A 21 s w   e e t #A    7
     |     |                         |                                 |
8   Num PL ;                         Np                                Vp                                    8
\end{BVerbatim}
}
\caption{A parsing via multiple alignment created by the SP computer model, like the one shown in Fig.~\ref{parsing_1_figure}, with the same sentence as before but with errors of omission, commission, and substitution as described in the text.}
\label{parsing_2_figure}
\end{figure*}

Fig.~\ref{parsing_2_figure} shows how the SP system may achieve a `correct' parsing of the same sentence as in Fig.~\ref{parsing_1_figure} but with errors: the addition of `\texttt{x}' within `\texttt{t h e}', the omission of `\texttt{l}' from `\texttt{a p p l e s}', and the substitution of `\texttt{k}' for `\texttt{w}' in `\texttt{s w e e t}'. In effect, the parsing identifies errors in the sentence and suggests corrections for them: `\texttt{t h x e}' should be `\texttt{t h e}', `\texttt{a p p e s}' should be `\texttt{a p p l e s}', and `\texttt{s k e e t}' should be `\texttt{s w e e t}'.

The system's ability to fill in gaps---such as the missing `\texttt{l}' in `\texttt{a p p l e s}'---is closely related to the system's ability to make probabilistic inferences---going beyond the information given---discussed in some detail in \cite[Chapter 7]{wolff_2006} and more briefly in \cite[Section 10]{sp_extended_overview}.

\subsection{Unsupervised Learning with Errors and Uncertainties in Data}\label{learning_with_errors_uncertainties_section}

Insights that have been achieved in research on language learning and grammatical inference \cite[Section 5.3]{sp_extended_overview}, \cite{wolff_1988}, \cite[Sections 2.2.12 and 12.6]{wolff_2006} may help to illuminate the problem of managing errors and uncertainties in big data.

\begin{figure}[!htbp]
\centering
\includegraphics[width=0.3\textwidth]{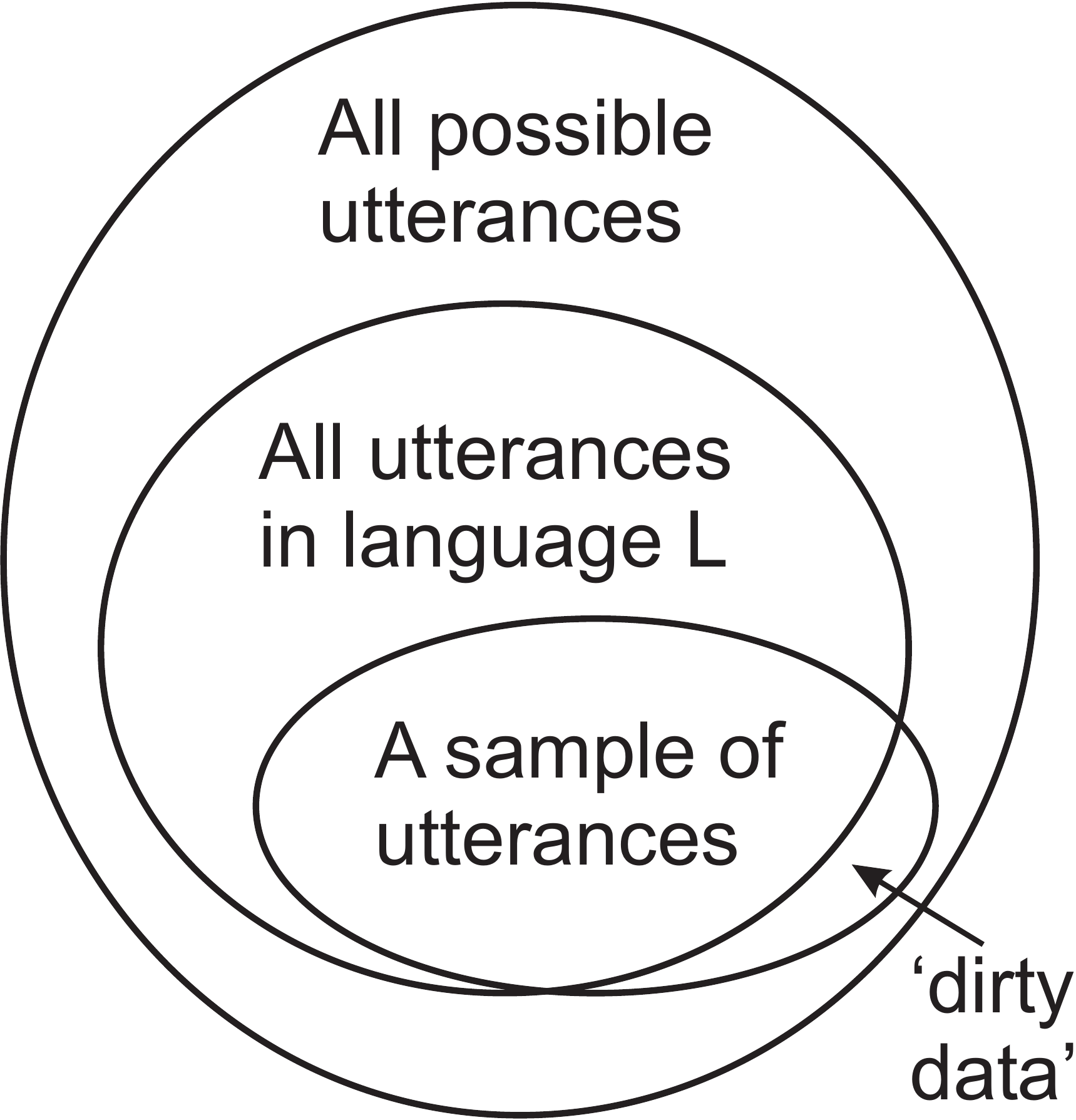}
\caption{Categories of utterances involved in the learning of a first language, $L$. In ascending order size, they are: the finite sample of utterances from which a child learns; the (infinite) set of utterances in $L$; and the (infinite) set of all possible utterances. Adapted from Fig.~7.1 in \cite{wolff_1988}, with permission.}
\label{generalisation_figure}
\end{figure}

The way we learn a first language has some key features:

\begin{itemize}

\item We learn from a finite sample of the language, normally quite large.\footnote{An alternative view, promoted most notably by Noam Chomsky, is that we are born with a knowledge of `universal grammar'---structures that appear in all the world's languages. But despite decades of research, there is still no satisfactory account of what that universal grammar may be or how it may function in the learning of a first language. Notice that the concept of a universal grammar is different from that of a UFK because the former means linguistic structures hypothesised to exist in all the world's languages, while the latter means a framework for the representation and processing of diverse kinds of knowledge.} This is represented by the smallest of the envelopes shown in Fig.~\ref{generalisation_figure}.

\item It is clear that mature knowledge of a given language, {\bf L}, includes an ability to interpret and, normally, to produce an infinite number of utterances in {\bf L}.\footnote{Exceptions in the latter case are people who can understand language but, because of physical handicap or other reason, may not be able to produce language (more below).} It also includes an ability to distinguish sharply between utterances that belong in {\bf L}---represented by the middle-sized envelope in Fig.~\ref{generalisation_figure}---and those that don't---represented by the area between the middle-sized envelope and the outer-most envelope in the figure.

\item The finite sample of language from which we learn includes many utterances which are not correct, meaning that they do not belong in {\bf L}. These include false starts, incomplete sentences, garbled words, and so on. These utterances are marked {\em dirty data} in the figure.

\end{itemize}

From these key features, two main questions arise, described here with putative answers provided by unsupervised learning via information compression:

\begin{itemize}

\item {\em Learning with dirty data}. How is it that we can develop a keen sense of what does or does not belong in our native language or languages, despite the fact that much of the speech that we hear as children contains the kinds of haphazard errors mentioned above, and in the face of evidence that language learning may be achieved without the benefit of error correction by a teacher, or anything equivalent.\footnote{In brief, the evidence is that people with a physical handicap that prevents them producing intelligible speech can still learn to understand their native language \cite{lenneberg_1962,brown_1989}. If such a child is saying nothing that is intelligible, there is nothing for adults to correct. Christy Brown \cite{brown_1989} went on to become a successful author, using his left foot for typing, and drawing on the knowledge of language that he learned by listening.}

    It appears that the principle of minimum length encoding (Section \ref{product_of_learning_section}) provides an answer.
    In a learning system that seeks to minimise the overall size of the `grammar' ({\bf G}) and the `encoding' ({\bf E}), most of the haphazard errors that people make in speaking---rare individually but collectively quite common---would be recorded largely in {\bf E}, leaving {\bf G} as a relatively clean expression of the language.

    Anything that is comparatively rare but exceeds the threshold for redundancy (Section \ref{product_of_learning_section}) may appear in {\bf G}, perhaps seen as a linguistic irregularity---such as `bought' (not `buyed') as the past tense of `buy'---or as a dialect form.

\item {\em Generalisation without over-generalisation}. How is it that, in learning a first language, {\bf L}, we can generalise from the finite sample of language which is the basis for learning to the infinite set of utterances that belongs in {\bf L}, without overgeneralising into the region between the middle-sized envelope and the outer-most envelope in Fig.~\ref{generalisation_figure}. As before, there is  evidence, discussed in the sources referenced above, that language learning does not depend on error-correction by a teacher or anything equivalent.

    As with learning with dirty data, it appears that generalisation without over-generalisation may be understood in terms of the principle of minimum length encoding. It appears that a learning process that seeks to minimise the overall size of {\bf G} and {\bf E} normally results in a grammar that generalises beyond the data in {\bf I} but does not over-generalise. Both under-generalisation and over-generalisation results in a greater overall size for {\bf G} and {\bf E}.

\end{itemize}

These principles apply to any kind of data, not just linguistic data. With unsupervised learning from a body of big data, {\bf I}, the SP system provides two broad options:

\begin{itemize}

\item Users may focus on both {\bf G} and {\bf E}, taking advantage of the system's capabilities in lossless information compression, and ignoring the system's potential with dirty data and the formation of generalisations without over-generalisation. This would be the best option in areas of application where the precise form of the data is important, including any `errors'.

\item By focussing on {\bf G} and ignoring {\bf E}, users may see the redundant features in {\bf I} and exclude everything else. As a rough generalisation, redundant features are likely to be `important'. They are likely to exclude most of the  haphazard errors in {\bf I} such as typos, misprints and other rarities that users may wish to ignore (but see Section \ref{rarity_probabilities_errors_section}). And {\bf G} is likely to generalise beyond what is in {\bf I}---filling in apparent gaps in the data---and to do so with generalisations that are sensitive to the statistical structure of {\bf I}, and excluding over-generalisations without that statistical support.

\end{itemize}

These two options are not mutually exclusive. Both would be available at all times, and users may adopt either or both of them according to need.

\subsection{Rarity, Probabilities, and Errors}\label{rarity_probabilities_errors_section}

Some issues relating to what has been said in Sections \ref{robust_against_errors_section} and \ref{learning_with_errors_uncertainties_section} are considered briefly here.

\subsubsection{Rarity and Interest}

It may seem odd to suggest that we might choose to ignore things that are rare, since antiques that are rare may attract great interest and command high prices, and conservationists often have a keen interest in animals or plants that are rare.

The key point here is that there is an important difference between a body of information to be mined for its recurrent structures and things like antiques, animals, or plants. The latter may be seen as information objects that are themselves the products of learning processes designed to extract redundancy from sensory data. Like other real-world objects, an antique chair is a persistent, recurrent feature of the world, and it is the recurrence of such an entity in different contexts that allows us to identify it as an object.

\subsubsection{The Flip Side of Probabilities}

As we have seen (Sections \ref{robust_against_errors_section} and \ref{learning_with_errors_uncertainties_section}), a probabilistic machine can help to identify probable errors in big data. But contradictory as it may seem, a consequence of working with probabilities---for both people and machines---is that mistakes may be made. We may bet on ``Desert King'' but find that ``Midnight Lady'' is the winner. And in the same way that people can be misled by a frequently-repeated lie, probabilistic machines are likely to be vulnerable to systematic distortions in data.

These observations may suggest that we should stick with computers in their traditional form, delivering precise, all-or-nothing answers. But:

\begin{itemize}

\item There are reasons to believe that computing and mathematics are fundamentally probabilistic: ``I have recently been able to take a further step along the path laid out by G{\"o}del and Turing. By translating a particular computer program into an algebraic equation of a type that was familiar even to the ancient Greeks, I have shown that there is randomness in the branch of pure mathematics known as number theory. My work indicates that---to borrow Einstein's metaphor---God sometimes plays dice with whole numbers.'' \cite[p. 80]{chaitin_1988}.

\item As noted in Section \ref{introduction_to_sp_section}, the SP system can be constrained to deliver all-or-nothing results in the manner of conventional computers. But ``constraint'' is the key word here: it appears that the comforting certainties of conventional computers come at the cost of restrictions in how they work, restrictions that may have been motivated originally by the low power of early computers \cite[p.~28]{wolff_2006}.

\end{itemize}

\section{Visualisation}\label{visualisation_section}

\begin{quote}

``... methods for visualization and exploration of complex and vast data constitute a crucial component of an analytics infrastructure.'' \cite[p.~133]{national_research_council_2013}.

\end{quote}

\begin{quote}

``[An area] that requires attention is the integration of visualization with statistical methods and other analytic techniques in order to support discovery and analysis.'' \cite[p.~142]{national_research_council_2013}.

\end{quote}

In the analysis of big data, it is likely to be helpful if the results of analysis, and analytic processes, can be displayed with static or moving images.

In this connection, the SP system has three main strengths:

\begin{itemize}

\item {\em Transparency in the representation of knowledge}. By contrast with sub-symbolic approaches to artificial intelligence, there is transparency in the representation of knowledge with SP patterns and their assembly into multiple alignments. Both SP patterns and multiple alignments may be displayed as they are or, where appropriate, translated into other graphical forms such as tree structures, networks, tables, plans, or chains of inference.

\item {\em Transparency in processing}. In building multiple alignments and deriving grammars and encodings, the SP system creates audit trails. These allow the processes to be inspected and could, with advantage, be displayed with moving images to show how knowledge structures are created.

\item {\em The DONSVIC principle}. As previously noted, the SP system aims to realise the DONSVIC principle \cite[Section 5.2]{sp_extended_overview} and is proving successful in that regard. This means that structures created or discovered by the system---entities, classes of entity, and so on---should be ones that people regard as natural. Those kinds of structures are also likely to be ones that are well suited to representation with static or moving images.

\end{itemize}

\section{A Road Map}\label{road_map_section}

\begin{figure}[!htbp]
\centering
\includegraphics[width=0.48\textwidth]{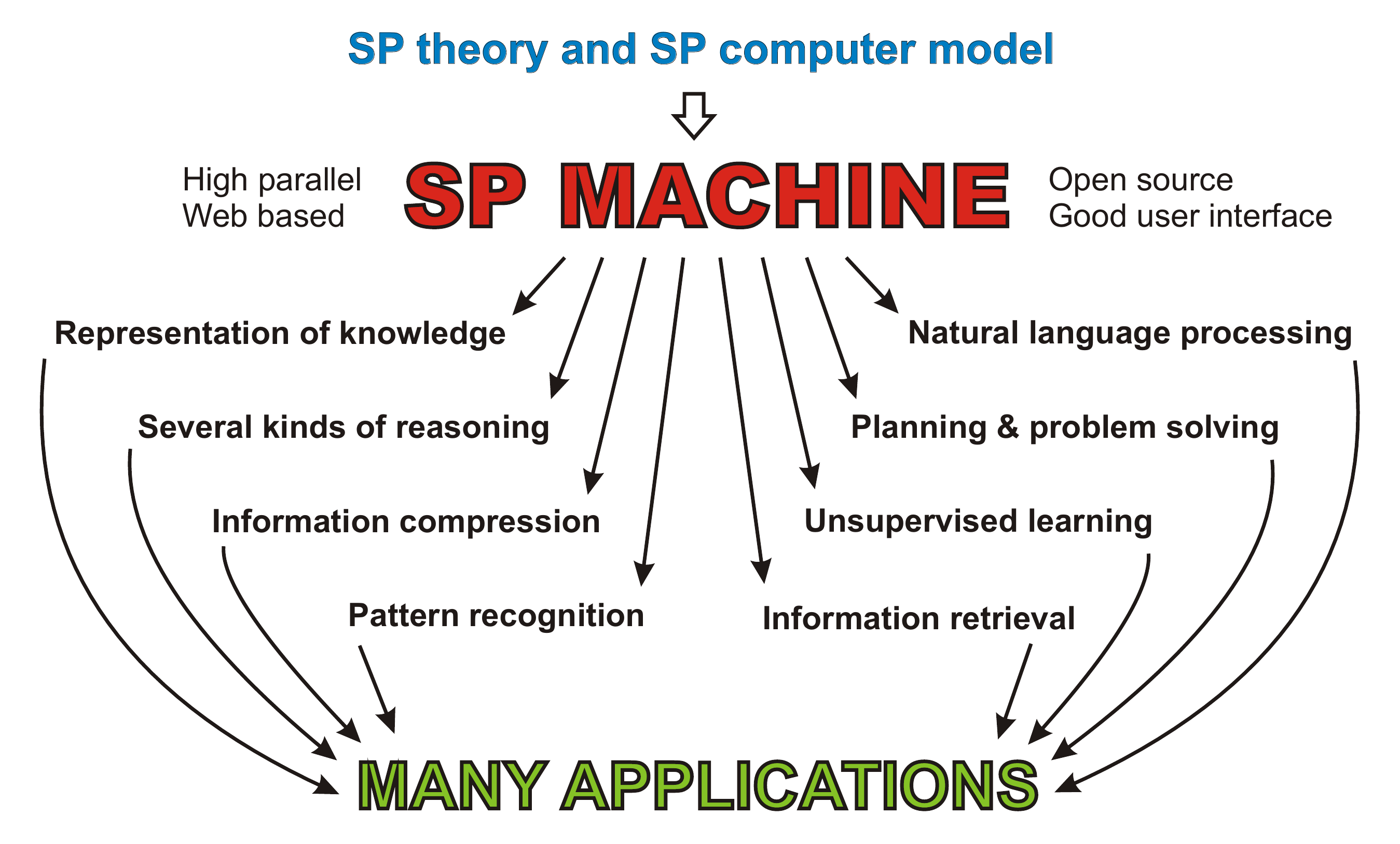}
\caption{Schematic representation of the development and application of the SP machine. Reproduced from Fig.~2 in \cite{sp_extended_overview}, with permission.}
\label{sp_machine_figure}
\end{figure}

As mentioned in Section \ref{introduction_to_sp_section}, it is envisaged that the SP computer model will provide the basis for the development of a new version of the SP machine. How things may develop is shown schematically in Fig.~\ref{sp_machine_figure}. It is envisaged that this new version will be realised as a software virtual machine, hosted on an existing high-performance computer, that it will employ high levels of parallelism, that it will be accessible via a user-friendly interface from anywhere in the world, that all software will be open source, and that users will be able to create new versions of the system. This high-parallel, open source version of the SP machine will be a means for researchers everywhere to explore what can be done with the system and to create new versions of it.

As argued persuasively in \cite[Chapters 5 and 6]{kelly_hamm_2013}, and echoed in this article in Sections \ref{potential_efficiency_ic_section} and \ref{data-centric_computing_section}, getting a proper grip on the problem of big data will probably require the development of new architectures for computers.

But there is plenty that can be done with existing computers. Most of the developments proposed in this article may be pursued without waiting for the development of new kinds of computer. Likewise, many of the potential benefits and applications of the SP system, described in \cite{sp_benefits_apps} and including such things as intelligent databases \cite{wolff_sp_intelligent_database} and new approaches to medical diagnosis \cite{wolff_medical_diagnosis}, may be realised with existing kinds of computer.

\section{Conclusion}

The SP system, designed to simplify and integrate concepts across artificial intelligence, mainstream computing, and human perception and cognition, has potential in the management and analysis of big data.

The SP system has potential as a {\em universal framework for the representation and processing of diverse kinds of knowledge} (UFK), helping to reduce the problem of variety in big data: the great diversity of formalisms and formats for knowledge, and how they are processed. The system may discover `natural' structures in big data, and it has strengths in the interpretation of data, including such things as pattern recognition, natural language processing, several kinds of reasoning, and more. It lends itself to the analysis of streaming data, helping to overcome the problem of velocity in big data.

Apart from several indirect benefits described in this article, information compression in the SP system is likely to yield direct benefits in the storage, management, and transmission of big data by making it smaller. The system has potential for substantial additional economies in the transmission of data (via the separation of encoding from grammar), and for substantial gains in computational efficiency (via information compression and probabilities, and via a synergy with data-centric computing), with consequent benefits in energy efficiency, greater speed of processing with a given computational resource, and reductions in the size and weight of computers. The system provides a handle on the problem of veracity in big data, with potential to assist in the management of errors and uncertainties in data. It may help, via static and moving images, in the visualisation of knowledge structures created by the system and in the visualisation of processes of discovery and interpretation.

The creation of a high-parallel, open-source version of the SP machine, as outlined in Section \ref{road_map_section}, would be a means for researchers everywhere to explore what can be done with the system and to create new versions of it.

\section{Acknowledgements}

I am grateful to Daniel J.~Wolff for drawing my attention to big data as an area where the SP system may make a contribution.

\bibliographystyle{IEEEtran}


\end{document}